\documentclass[letterpaper, 10 pt, conference]{ieeeconf}  

\IEEEoverridecommandlockouts                              
\usepackage{graphics} 
\usepackage{epsfig} 
\usepackage{mathptmx} 
\usepackage{times} 
\usepackage{amsmath} 
\usepackage{amssymb}  
\usepackage{mathtools}

\usepackage[noadjust]{cite}

\newcommand{\norm}[1]{\left\lVert#1\right\rVert}
\newcommand{\etal}{\textit{et~al}.~}

\title{\LARGE \bf
Collaborative Robot Learning from Demonstrations \\ using Hidden Markov Model State Distribution
}

\author{Sulabh Kumra and Ferat Sahin
\thanks{Sulabh Kumra and Ferat Sahin are with the Department of Electrical Engineering, Rochester Institute of Technology, Rochester, NY, USA.
        {\tt\small sk2881@rit.edu, feseee@rit.edu}}%
}

\begin{document}

\maketitle
\thispagestyle{empty}
\pagestyle{empty}

\begin{abstract}

In robotics, there is need of an interactive and expedite learning method as experience is expensive. Robot Learning from Demonstration (RLfD) enables a robot to learn a policy from demonstrations performed by teacher. RLfD enables a human user to add new capabilities to a robot in an intuitive manner, without explicitly reprogramming it. In this work, we present a novel interactive framework, where a collaborative robot learns skills for trajectory based tasks from demonstrations performed by a human teacher. The robot extracts features from each demonstration called as key-points and learns a model of the demonstrated skill using Hidden Markov Model (HMM). Our experimental results show that the learned model can be used to produce a generalized trajectory based skill.

\end{abstract}

\section{INTRODUCTION}

Nowadays, most of the robots used in the industry are pre-programmed and necessitate a well-defined and controlled environment. Reprogramming these robots is often an expensive process necessitating an expert. Enabling the robot to learn tasks by demonstrating them would streamline the robot installation and task reprogramming. That is why, Robot Learning from Demonstrations (RLfD) \cite{Atkeson1997RobotDemonstration} is one of the key research areas in the field of robotics. However, constructing a robot that is able to learn by observation is still a challenging problem. Although, prototype platforms for robot learning by demonstration have been around for more than a decade, many complications have restrained the robots to operate only in restricted laboratory environments. Some of the key challenges are perception, task recognition, task generalization, motion planning, and object manipulation.

RLfD refers to the technique of teaching skills to a robot by giving examples of the desired behavior through human demonstrations. It is similar to the way human beings learn a new skill from demonstrations performed by the teacher. One of the great advantage of this technique is that it eliminates the need of expert level technical knowledge to program a robot. Moreover, it captures key features from the demonstration provided by a teacher who is expert in the specific task, which a robotic programmer might not be able to program. Thus, this technique has great potential in industrial as well as home robotic applications.

Learning the mapping between world state and actions is called as a \textit{policy}. This allows a robot to choose an action based on its current state. In RLfD, a robot learns a new policy from demonstrations provided by the human teacher. A demonstration is defined as sequence of state–action pairs that are recorded throughout the performance of the required robot behavior by a human teacher. RLfD uses the dataset of recorded demonstrations to form policy to reproduce the demonstrated behavior.

We focus on learning, representing, and generalizing of trajectory based skills from demonstrations. We propose a novel RLfD framework by which a collaborative robot can learn trajectory based skills from kinesthetic demonstrations performed by a human teacher. The prosed RLfD framework has three fundamental phases: data collection from demonstrations, learning behavior by deriving a policy, and robot execution or reproduction of the demonstrated behavior. It extracts key information from multiple demonstrations to learn a state-action policy, which is used to produce a generalized trajectory based skill.

\begin{figure}
\begin{center}
\includegraphics[width=0.85\linewidth]{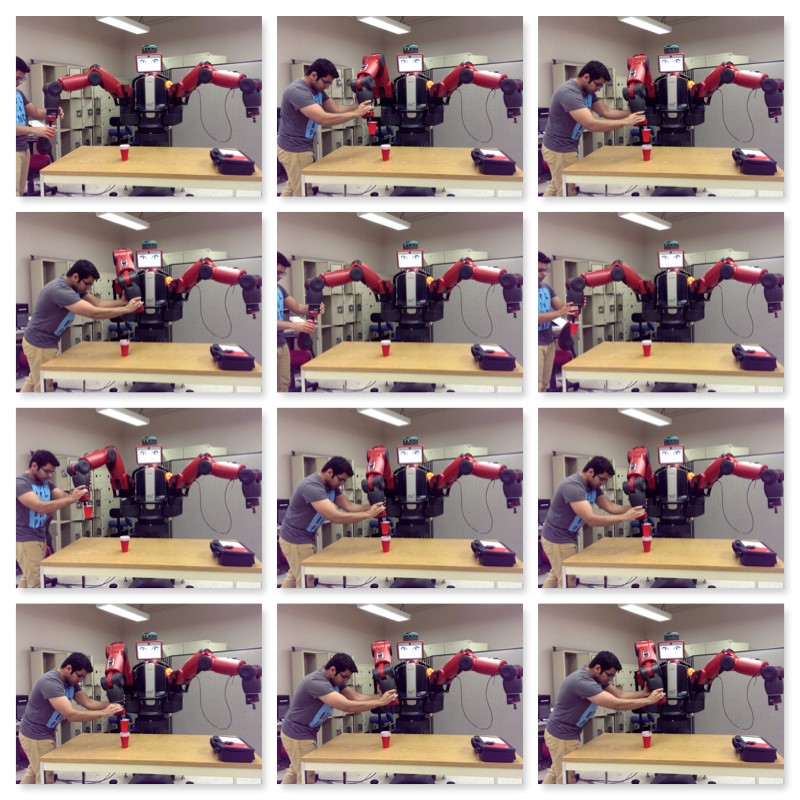}
\end{center}
   \caption{Human performing demonstration using kinesthetic teaching.}
\label{fig:demo}
\end{figure}


\section{BACKGROUND}

RLfD is a technique to enable a robot to perform new tasks autonomously. Instead of necessitating users to logically decompose and manually program a robot for a desired behavior, RLfD enables to derive robot controller by observing human's own performance. The goal is to easily extend and adapt robot capabilities to novel situations, even by users who lag programming ability. Robot learns a model of a task based on the demonstrations performed by the teacher. In the last two decades, several RLfD approaches have been developed \cite{Argall2009ADemonstration}. However, until date most of the robots are tediously hand programmed for each of the task they have to perform. RLfD tries to minimize this challenging step by allowing users to teach their robot to fit in their specific needs. RLfD techniques are user friendly and enable robots to be employed more widely in daily activities with no need of specialist humans. Moreover, by using expert knowledge of human teacher, in form of demonstrations, the overall learning process would be faster as compared to hand-engineered control policy and learning by trial-and-error.

One of the well-known works in RLfD is by Calinon \etal \cite{Calinon2007OnRobot} \cite{Calinon2010Learning-basedRedundancies} \cite{Calinon2015RobotModels}, in which a probabilistic representation of the demonstrations is built using Gaussian Mixture Model (GMM) and a smooth trajectory is reproduced using Gaussian Mixture Regression. In another work by Schneider \etal introduced a Gaussian process regression model that clusters the input space from performed demonstrations into smaller subsets and work individually on these subsets \cite{Schneider2010RobotRegression}. Both of these Gaussian techniques generalized over a set of demonstrated trajectories, but the GMM is computationally expensive.

Rozo \etal proposed an end-to-end RLfD framework for teaching force-based manipulation tasks \cite{Rozo2013ATasks}. The demonstrations are encoded using a Hidden Markov Model (HMM), and the reproduction is done using a modified version of Gaussian Mixture Regression. Asfour \etal used continuous HMM to generalize movements demonstrated to a robot multiple times. The experiments were performed using a kinematics model of the human upper body to simulate the reproduction of arm movements \cite{Asfour2008ImitationRobots}.

In a recent work, Ahmadzadeh \etal proposed an approach that generates a continuous representation of the demonstrated trajectory based on the concept of canal surfaces \cite{Ahmadzadeh2016TrajectoryApproach}. Their experimental results show that the proposed approach can reproduce a wide range of trajectories that achieve the goal of the task. Koskinopoulou \etal developed a framework for Human Robot Collaborative (HRC) task execution \cite{Koskinopoulou2016LearningExecution}. Akgun and Thomaz developed an algorithm to simultaneously learning actions and goals from naïve human teacher demonstrations \cite{Akgun2016SimultaneouslyDemonstration}.

Konidaris \etal presented an online algorithm for constructing skill trees from demonstration trajectories in a dynamic continuous domain \cite{Konidaris2012RobotTrees}. They evaluated their algorithm on the uBot-5 mobile manipulator and showed that it was able to learn skills from both expert demonstration and learned control
sequences. Other techniques like  linguistic transfer of an
assembly task from human to robot have also been developed \cite{Dantam2012LinguisticRobots}.


\section{PROBLEM STATEMENT}

\begin{figure}
\begin{center}
\includegraphics[width=0.7\linewidth]{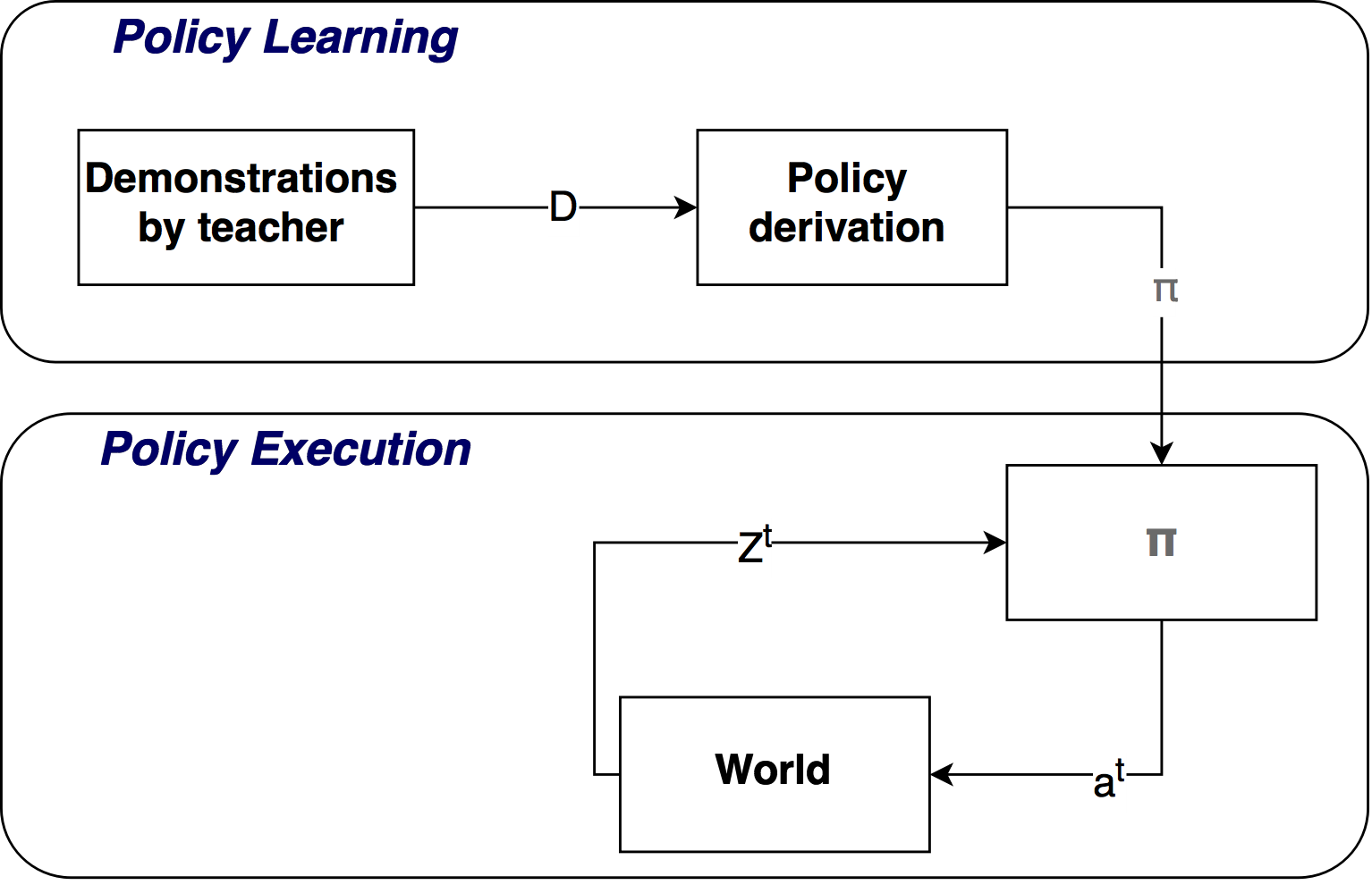}
\end{center}
   \caption{Control policy learning and execution.}
\label{fig:problem}
\end{figure}

We can consider RLfD as a subset of Supervised Learning. In supervised learning, the agent takes the labelled training data and the algorithm learns an approximate model to fit this data. Similarly, in RLfD, the training data is collected from demonstrations provided by the teacher. Figure \ref{fig:problem} (top) shows training data $D$ being acquired from demonstrations by teacher to derive a policy. 

The robot world comprises of states $S$ and actions $A$. Mapping between the states, is defined by a probabilistic transition function called state transition matrix given by:

\begin{equation}
T(s\prime \vert s,a): S \times A \times S \rightarrow [0,1]  
\end{equation}

The set A contains high-level behaviors as well as low-level motions. We make an assumption that the states are not fully-observable or hidden. However, the learning algorithm has access to the observed state $Z$, with the mapping $M: S \rightarrow Z$. The policy $\pi: Z \rightarrow A$ selects the next action based on the current robot state. Figure \ref{fig:problem} (bottom) shows one cycle of policy execution.

\section{PRPOSED APPROACH}

\begin{figure*}
\begin{center}
\includegraphics[width=1\linewidth]{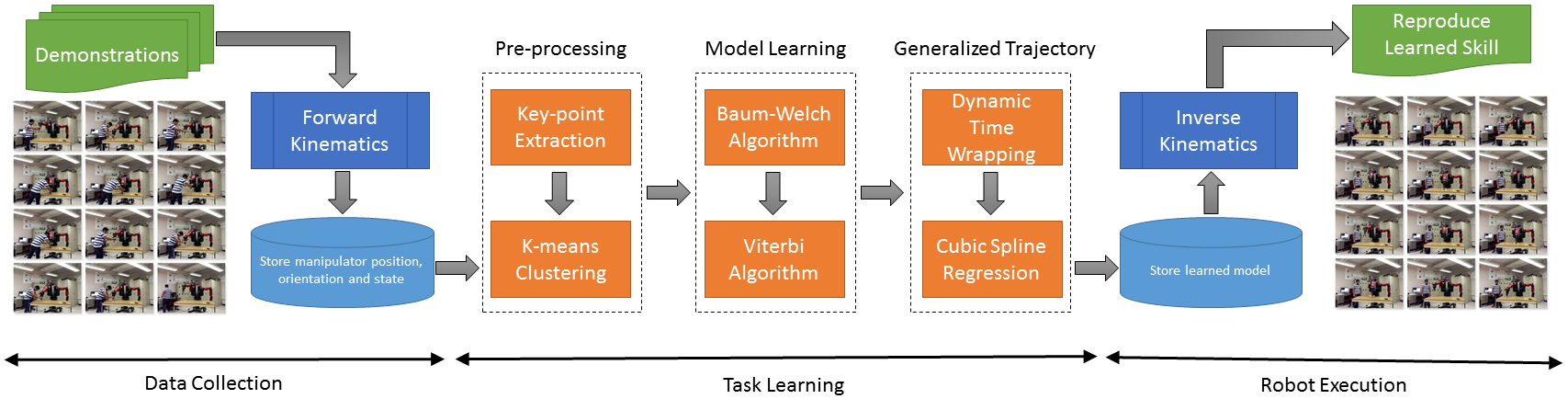}
\end{center}
   \caption{Block diagram of the proposed RLfD system.}
\label{fig:approach}
\end{figure*}

The proposed RLfD system consists of three fundamental phases. Figure \ref{fig:approach} shows a block diagram of the proposed approach. First phase is the data collection phase, in which the data is acquired from the demonstrations performed by the teacher. Real-time joint angle values and gripper state are received by the system, which applies Forward Kinematics to find the position and orientation of the manipulator. This process is repeated for all the demonstrations and the time series of position and orientation are stored.

The second phase is the task learning phase, in which the stored data from the first phase is used to learn a model of the task or the behavior. From each demonstration, key-points of trajectories are extracted and then \textit{k}-means clustering is used to cluster the key-points from all demonstrations. Centers of each cluster is calculated and mapped to a unique symbol representing a state. Baum-Welch algorithm is used to learn a policy from the demonstrations to find the probability matrix of state-action pairs. Then, Viterbi algorithm is used to find the most probable sequence of states. Dynamic Time Warping (DTW) is used to align the time vector of the learned trajectory.

Third phase is the robot execution phase, in which the saved model is used to produce the learned behavior. Cubic smoothing-spline regression is used to determine a generalized trajectory from the most probable sequence of states. Inverse kinematics is applied to obtain a generalized trajectory suitable for the robot.

\section{DATA COLLECTION}

Data collection is the first phase of the system, where the robot collects data from the physical demonstrations performed by human teacher. In the next three sub-sections, we discuss the main steps of this phase.

\subsection{Demonstrations}

In our work, we have used kinesthetic teaching to demonstrate the task or trajectories to the robot. Kinesthetic teaching is a way of teaching in which, the teacher physically holds the robot arm or manipulator, and demonstrates the task by moving the robotic arm. The main advantage of this technique is that there is no need to map the real world space to robot world space, as the performed demonstrations are with respect to robot world.

One major roadblock for this type of teaching is that it requires a robot that has a feedback system to record the joint position in real-time. However, we do not face this problem as we are using one of the most advanced research robot called Baxter, which has inbuilt feedback system to measure the joint angles from all fourteen joints and manipulator state of the dual-arm robot. This allows us to read these joint values from joint trajectory server in real time. To get more precision for fast movements, we have increased the default rate (100Hz) of reading current joint angles from the robot to 1000 Hz. These values are stored with a timestamp and sent to the forward kinematics function.

Figure \ref{fig:demo} shows an example demonstration being performed using kinesthetic teaching to demonstrate a pick and place task to Baxter. In this, the teacher holds the force sensor at the manipulator of each robotic arm and guides the robot through the task. To change the state of the gripper, physical buttons near the manipulator are used. One button closes the gripper and the other opens the gripper. The system records the gripper states along with the joint angles throughout the demonstration, and for each demonstration.

\subsection{Forward Kinematics}

‘Forward kinematics refers to the use of the kinematic equations of a robot to compute the position of the end-effector from specified values for the joint parameters’. Kinematics equations are series of transformation matrices to characterize the relative movement at each joint with respect to the previous joint. Instead of going through the complex matrix multiplications for calculating the relative movements at each joint, we use Denavit-Hartenberg parameters or DH parameters.

Each individual homogeneous transformation $A_i$ is calculated as:

\begin{equation}
\begin{aligned}
A_i &= Rot_{z,\theta_i}Trans_{z,d_i}Trans_{x,a_i}Rot_{x,a_i} \\
&= \begin{bmatrix}C_{\theta_i}&-S_{\theta_i}C_{a_i}&S_{\theta_i}S_{a_i}&\alpha_iC_{\theta_i}\\S_{\theta_i}&C_{\theta_i}C_{a_i}&-C_{\theta_i}S_{a_i}&\alpha_iS_{\theta_i}\\0&S_{a_i}&C_{a_i}&0\\0&0&0&1\end{bmatrix}
\end{aligned}
\end{equation}

\noindent where $\theta_i, d_i, \alpha_i$ and $a_i$ are known as DH parameters associated with link $i$ and joint $i$.

\begin{figure}
\begin{center}
\includegraphics[width=0.65\linewidth]{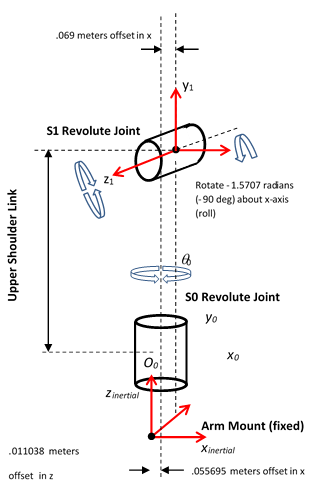}
\end{center}
   \caption{Approach to mathematically model Baxter’s arm}
\label{fig:arm_model}
\end{figure}

We use Baxter as our robot for learning, thus we calculated the DH parameters for both the limbs of the robot to perform the forward kinematics. We follow the standard convention of placing the frames at each joint to calculate the DH table. Figure \ref{fig:arm_model} shows an example of placing the frames on first two revolute joints and the physical parameters used to mathematically modeling Baxter’s arm. Table \ref{dh_table} shows the derived DH parameters for all the joints and links for Baxter. These parameters are used to calculate the individual transformation matrices for each of the link.

\begin{table}
\centering
\caption{DH Table for Baxter}
\label{dh_table}
\begin{tabular}{|l|l|l|l|l|}
\hline
Joint & $\theta$ & $d$      & $a$     & $\alpha$    \\
\hline
1     & $\theta_1$ & 0.2703 & 0.069 & -1.571 \\
2     & $\theta_2$ & 0      & 0     & 1.571  \\
3     & $\theta_3$ & 0.3644 & 0.069 & -1.571 \\
4     & $\theta_4$ & 0      & 0     & 1.571  \\
5     & $\theta_5$ & 0.3743 & 0.01  & -1.571 \\
6     & $\theta_6$ & 0      & 0     & 1.571  \\
7     & $\theta_7$ & 0.2295 & 0     & 0      \\
\hline
\end{tabular}
\end{table}

Now, we use the DH parameters to calculate $A_1, A_2, .... A_7$. The homogeneous transformation matrix that expresses the position and orientation of the manipulator is called a transformation matrix $T$, and is calculated as:

\begin{equation}
T = A_1 \times A_2 \times A_3 \times A_4 \times A_5 \times A_6 \times A_7  
\end{equation}

For each measurement, the calculated position and orientation are stored into a data file, along with the time stamp and the gripper state. This marks the end of the first phase.


\section{TASK LEARNING}

Task learning is the second phase of the system. In this phase, the data stored during the first phase is used to learn a model of the task.

\subsection{Key-point Extraction}
\label{kp}
\begin{figure}
\begin{center}
\includegraphics[width=0.5\linewidth]{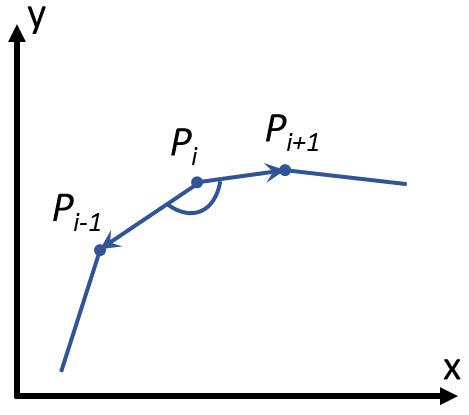}
\end{center}
   \caption{Key-point selection criteria}
\label{fig:key_point}
\end{figure}

We detect the characteristic features of each demonstration, and we call it key-points. By doing this, we avoid having large number of hidden states to be trained by HMM. We represent every recorded movement during the demonstration by a set of time-discrete sequence. For each arm we have a sequence $P_{d,1},P_{d,2},...,P_{d,len(d)}$ that defines the positions of manipulator, and a sequence $O_{d,1},O_{d,2},...,O_{d,len(d)}$ that defines the orientations of manipulator over time, where $l(d)$ is defined as the length of demonstration $d$. $P_i$ is a three dimensional vector, whereas $O_i$ is a four dimensional vector. A point $P_{d,i}$ is selected as a key-point $K_{d,j}$ in the sequence of points if: 

\begin{equation}
\begin{aligned}
& \angle ( \overrightarrow{P_{d,t}} - \overrightarrow{P_{d,t-1}}, \overrightarrow{P_{d,t}}, \overrightarrow{P_{d,t}} - \overrightarrow{P_{d,t+1}}) < 2\pi - \epsilon_1   \\
\vee~~ & \norm{\overrightarrow{P_{d,t}} - \overrightarrow{P_{d,t-1}}} < \epsilon_2, i - \tau_{d,j-1}>\epsilon_3, \\
& \norm{\overrightarrow{P_{d,t}} - \overrightarrow{P_{d,\tau_{d,j-1}}}} > \epsilon_4 \\
\vee~~ & \norm{\overrightarrow{P_{d,t}} - \overrightarrow{P_{d,\tau_{d,j-1}}}} \geq \epsilon_5, i - \tau_{d,j-1}>\epsilon_6, \\
& \norm{\overrightarrow{P_{d,n}} - \overrightarrow{P_{d,\tau_{d,j-1}}}} < \epsilon_7 ~ \forall ~ n \in  [\tau_{d,j-1}, i)
\end{aligned}
\end{equation}

\noindent where, $\tau_{d,j}$ denotes the time-stamp of demonstration $d$ and $j^{th}$ key-point. This means that if the angle between the vector that goes from point $P_i$ to its successor $P_{i+1}$ and the vector that goes from point $P_i$ to its predecessor $P_{i-1}$ is less than $2\pi - \epsilon_1$, then point $P_i$ is selected as a key-point. Figure \ref{fig:key_point} shows an example of finding angles between the two vectors in a two dimensional plane. To detect only sharp corners in the manipulator trajectory as key-points $\epsilon_1$ should be high.

To detect the key-points $O_{d,j}$ of the orientation angles of the manipulator, similar selection criterion for key-points is used. A point $O_d$,i is selected as a key-point $K_{d,j}$ in the sequence of points $O_{d,1},O_{d,2},...,O_{d,len(d)}$ if: 

\begin{equation}
\begin{aligned}
& \angle ( \overrightarrow{O_{d,t}} - \overrightarrow{O_{d,t-1}}, \overrightarrow{O_{d,t}}, \overrightarrow{O_{d,t}} - \overrightarrow{O_{d,t+1}}) < 2\pi - \epsilon_8   \\
\vee~~ & \norm{\overrightarrow{O_{d,t}} - \overrightarrow{O_{d,t-1}}} < \epsilon_9, i - \tau_{d,j-1}>\epsilon_{10}, \\
& \norm{\overrightarrow{O_{d,t}} - \overrightarrow{O_{d,\tau_{d,j-1}}}} > \epsilon_{11} \\
\vee~~ & \norm{\overrightarrow{O_{d,t}} - \overrightarrow{O_{d,\tau_{d,j-1}}}} \geq \epsilon_{12}, i - \tau_{d,j-1}>\epsilon_{13}, \\
& \norm{\overrightarrow{O_{d,n}} - \overrightarrow{O_{d,\tau_{d,j-1}}}} < \epsilon_{14} ~ \forall ~ n \in  [\tau_{d,j-1}, i)
\end{aligned}
\end{equation}

A major challenge in detection of the key-points using this approach is to tune the values of thresholds $\epsilon_1 - \epsilon_{14}$, which will decide the number of key-points extracted from the recorded task or trajectories. If the number of detected key-points is low, some relevant characteristics of the task or trajectories can be missed from the generalization step. Whereas, if the detection of key-points is high, over fitting of the recorded task or trajectories can take place.

\begin{figure}
\begin{center}
\includegraphics[width=1\linewidth]{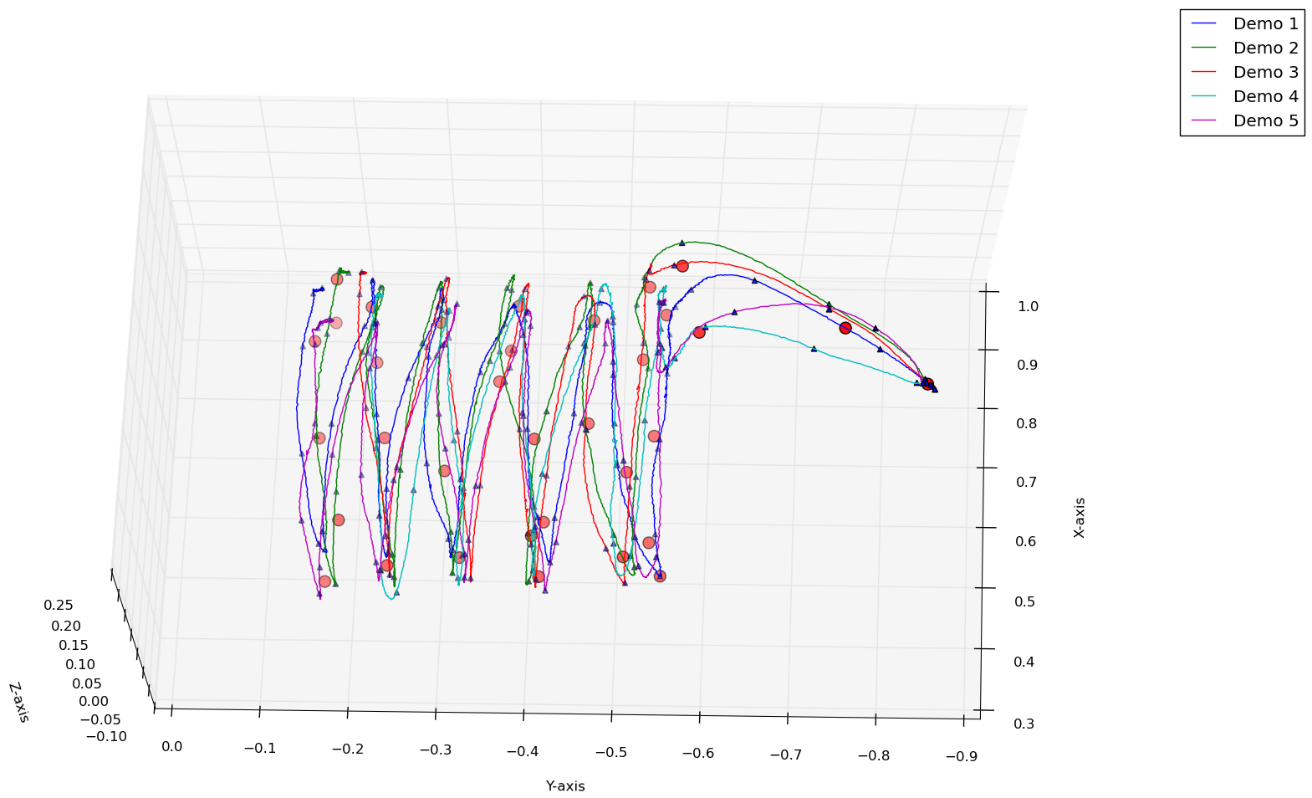}
\end{center}
   \caption{The small points are the extracted key-points from multiple demonstrations and the big points are the centroids of the k clusters}
\label{fig:k_cluster}
\end{figure}

Keeping in mind the trade-off between generalization and over-fitting, reasonable values for $\epsilon_1 - \epsilon_{14}$ were experimentally determined. Figure \ref{fig:k_cluster} shows the result of key-point extraction algorithm on multiple demonstrations. It shows the key-points extracted from three demonstrations. The key-points are marked with circles on each original trajectory of the demonstration.

\subsection{Clustering}
\label{clust}
In contrast to the common key point approach used by Asfour \etal \cite{Asfour2008ImitationRobots}, we cluster the key-points derived from all the demonstrations and calculated the centroids of each cluster. To solve this problem, we use \textit{k}-means clustering algorithm. The \textit{k}-means clustering algorithm takes the number of clusters as an input to generate \textit{k} clusters, from a set of observation vectors. It returns the centroids for each of the \textit{k} clusters formed. For our system, we set the number of clusters \textit{k} equal to average number of key-point in each demonstration.

A key-point vector is classified with a cluster if the centroid of the cluster is closest to it, i.e. closer to centroid than any other centroids. \textit{k}-means clustering algorithm attempts to minimize the distortion, which is the sum of the squared distances between each key-point vector and its dominating centroid. At each step, \textit{k}-means refines the choice of centroids and tries to reduce the distortion. When this distortion change gets below a threshold, the algorithm stops. Figure \ref{fig:k_cluster} shows the result of the implement \textit{k}-means clustering algorithm to cluster the key-points and find the centroids of the clusters. In the plot, each centroid is marked with big red circles. In the next section, we discuss how these centroids are used to learn a model of the task or trajectories using a HMM.

\subsection{Model Learning using HMM}

A HMM (Hidden Markov Model) is a directed graphical model that is a statistical Markov chain of a sequence of unobserved or hidden states and a corresponding sequence of observation variables. It has been widely applied in the field of handwriting recognition, speech recognition, DNA sequence analysis, etc.

\begin{figure}[ht]
\begin{center}
\includegraphics[width=0.7\linewidth]{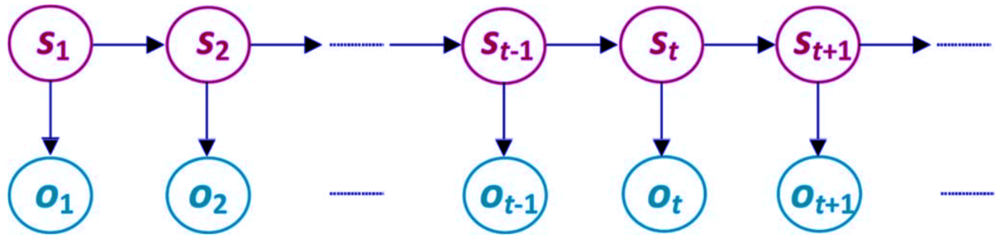}
\end{center}
   \caption{Graphical representation of HMM}
\label{fig:hmm}
\end{figure}

Figure \ref{fig:hmm} shows a graphical representation of HMM, where $S_t$ denotes the hidden states and $O_t$ denotes the observed variables at time instants $t \in \{1, 2, 3,.. , t-1, t, t+1, ... \}$. The probability of going to $j^{th}$ state at time $t+1$, given that the current is state $i$ at time $t$, is denoted by:

\begin{equation}
a_{ij} = P(S_{t+1} = j ~\vert~ S_t = i)
\end{equation}

Using these probabilities, we form the state transition matrix, given by:

\begin{equation}
A = \{a_{ij}\}, ~\forall~ i,j = \{1,2,3, ..., N_s\}
\end{equation}

\noindent where, $N_s$ denotes the number of hidden states in the HMM. In this work, the number of hidden states $N_s$ was set equal to the number of centroids derived from the demonstrations.

The initial state probabilities is defined as the probability of model being in state $i$ at time $t = 1$, and is given by:

\begin{equation}
\pi = \{\pi_i = P(S_1=i)\}, ~\forall~ i = \{1,2,3, ..., N_s\}
\end{equation}

The observation probability is defined as the probability of observing a symbol $q_k$ at time $t$ given the model is in state $i$, and is denoted as:

\begin{equation}
b_i(k) = P(q_k ~\vert~ S_t = i)
\end{equation}

Using these probabilities, we form the observation probability matrix, given by:

\begin{equation}
B = \{b_i(k)\}, ~\forall~ i = \{1,2,3, ..., N_s\} and k = \{1,2,3, ..., Q\}
\end{equation}

\noindent where, $Q$ is the number of observation symbols. A complete HMM is described as:

\begin{equation}
\lambda = \{\pi, A, B\}
\end{equation}

In our research, we used a discrete-HMM for learning a model of the demonstrated trajectories, which required the recorded continuous trajectories to be mapped to discrete values. The key-point extraction and clustering technique applied in section \ref{kp} Key-point Extraction and section \ref{clust} Clustering were for vector quantization. As it is required to use most of the k clusters in order to preserve most of the features in the continuous trajectory, we mapped each cluster into a discrete symbol $o_{n,m}$ to from a codebook of symbols $\{q_1, q_2, q_3, ...., q_k\}$, where k is the number of clusters formed. The sequences of these observations form the set, and is given by:

\begin{equation}
\textbf{O} = \{O_m = (o_{1,m}, o_{2,m}, o_{3,m}, ..., o_{N_m,m})\}, ~\forall~ m = \{1,2,3, ..., M\}
\end{equation}

\noindent where, M denotes the total number of demonstrations.

The efficiency of learning with HMM depends on the number of available observations. As in RLfD, it is preferred to keep the number of observations or demonstrations low, an appropriate initialization of the model parameters is imperative. Next, we discuss the initialization process of the HMM parameters.

We implemented the Bakis left–right topology \cite{Rabiner1989ARecognition} to model the demonstrated task. The forward-transition probabilities $(\alpha_{i,i+1}, \alpha_{i,i+2}, ...)$ and self-transition probabilities $(\alpha_{i,i})$ were initialized in the state transition matrix $A_{i,j}$ as:

\begin{equation}
\begin{aligned}
& \alpha_{i, i+1} = (\frac{1}{\tau_{i,\delta}})(\frac{1}{Z}), \\
& \alpha_{i, i+2} = (\frac{1}{4\tau_{i,\delta}})(\frac{1}{Z}), \\
and ~~& \alpha_{i, i} = (1-\frac{1}{\tau_{i,\delta}})(\frac{1}{Z})
\end{aligned}
\end{equation}

\noindent where, $\tau_{i,\delta}$ is the amount of time spent in state $i$, and
$Z$ is a normalizing constant to make sure $ \sum \alpha_(i,j)= 1$. 

All other state transition probabilities were set as zero, to make sure that transition to those states is impossible. Output probabilities $b_i (k)$ were initialized as:

\begin{equation}
b_i (k) = n_{i,\delta} (q_k)/\tau_{i,\delta}
\end{equation}

\noindent where, $n_{i,\delta} (q_k)$ denotes the number of times $q_k$ is observed in state $i$ of $O_\delta$. The state probabilities $\pi$, were initialized as $\pi = [1~ 0~...~0]$. This ensures that the learning always starts from the starting point.

Once the model parameters were initialized, Baum–Welch algorithm \cite{Rabiner1989ARecognition} was used to train on all demonstrations or observations $O_1, O_2, ... , O_M$. Later, Viterbi algorithm \cite{Rabiner1989ARecognition} was used to determine the most probable sequence of hidden state. Due to the nature of the implemented Bakis left-right topology, some hidden states or centroid points were not present in all the observed trajectories. We called them as zero-points and were ignored in the generalized trajectory.

\subsection{Dynamic Time Warping}

Since the length and velocities of the demonstrated trajectories differ, the time frames of extracted key-points of each demonstration are different. To produce the learned trajectory, we need to align the set of key-points along a common time vector. To solve this problem, we use DTW \cite{Keogh2005ExactWarping}, in which the temporal alignment of the clusters made by key-points is done by aligning the entire trajectory with respect to a reference trajectory.

DTW sequence alignment technique forms a matrix that consists of distances between two time-series, which is then used to find an optimal path that minimizes the distance between the two time-series. For a given test sequence $t_1, t_2, ..., t_T$ of length $T$, and a reference sequence $r_1, r_2, ..., r_R$ of length $R$, the distance matrix is calculated as:

\begin{equation}
H(x,y)= \norm{r_x-t_y}^2, ~\forall~  x=1,2,..,R, ~and~ y=1,2,..,T
\end{equation}

In our research, to calculate the distance between two time-`series, Euclidean l2-norm is used. The optimal alignment path $g(x,y)$  is calculated as:

\begin{equation}
g(x,y) = H(x,y) + min \{g(x-1,y),g(x-1,y-1),g(x,y-1)\}
\end{equation}

For selection of the reference sequence, forward algorithm was used to find the demonstration that has the highest probability for the learned model. As our data received from demonstration was eight dimensional data, we implemented multidimensional DTW, and the distance matrix is calculated as:

\begin{equation}
H(x,y)= \sum_{d=1}^D\norm{r_x^d-t_y^d}^2 ,  \forall  x = 1,...,R,~and~ y = 1,...,T
\end{equation}

\noindent where, $D$ is the number of dimensions, and in our case, $D = 8$.

\subsection{Generalized Trajectory}

After we have a common time-line for the model learned for the sequence of centroids, we need to connect these centroids to generate a generalized trajectory. To solve this problem, we used cubic smoothing-spline regression to determine a generalized trajectory from the most probable sequence of centroids. This method is widely applied for fitting a smooth curve to large set of scattered data.  In our case, we want to generate a smooth trajectory from the set of scattered centroid points. The spline curve was interpolated at intervals equal to the period size of clusters, which produced a generalized trajectory suitable for Baxter.

\section{ROBOT EXECUTION}

Once the model of the task is saved, we use the stored files to perform the task. The files contains the centroids, which are the position and orientation of the manipulator. For the robot to reproduce the learned task, we need to convert these position and orientation values into joint angles values that are suitable for the robot. To solve this problem, we use Inverse Kinematics. This technique is just the opposite of Forward Kinematics. It provides the set of joint angle values for a given the Cartesian pose of the manipulator.

We use the Baxter’s IK server to perform the inverse kinematics during the execution of the task. It is a build-in service from Baxter SDK. It takes the position and orientation values as the input and returns seven valid joint angles values to achieve it. These values are sent to the robot and it moves to that state using Joint Trajectory Action Server (JTAS).

\section{EXPERIMENTS}

To evaluate the approach, we performed the experiments on the research robot by Rethink Robotics called Baxter. It is a dual-arm 7-DoF robot. A modular approach was used to create the framework and all the algorithms were developed as separate packages in python. This made it easier to test and debug each part of the framework separately. In this section, we discuss experiments performed and the results of those experiments.

The key-point extraction algorithm and clustering algorithm plays a vital role in the efficiency of the model. The variables for tuning the key-point extraction algorithm were experimentally determined. To make sure this algorithm works well on wide variety of task and complex trajectories, we validated them on different tasks. In the next two sub-sections, we discuss the details of the two complex tasks that the robot learned through multiple demonstrations.

\subsection{Stacking cup task}

The first task is to place a cup on a stack of cups. In this task, the robot takes the cup from a human and puts it on a stack of cups. Figure \ref{fig:demo} shows robot learning the stacking cup task from a human teacher through demonstration. Three demonstrations were performed to teach this task. Table \ref{data_cup} shows the data statistics for this task.

\begin{table}
\centering
\caption{Data statistics for stacking cup task}
\label{data_cup}
\begin{tabular}{|c|c|c|}
\hline
Demonstration \# & \# of recorded values & \# of key-points extracted \\
\hline
Demonstration  1     & 18024 & 18  \\
Demonstration  2     & 19759 & 19   \\
Demonstration 3      & 18140 & 20 \\
\hline
\end{tabular}
\end{table}

\begin{figure}[h]
\begin{center}
\includegraphics[width=0.8\linewidth]{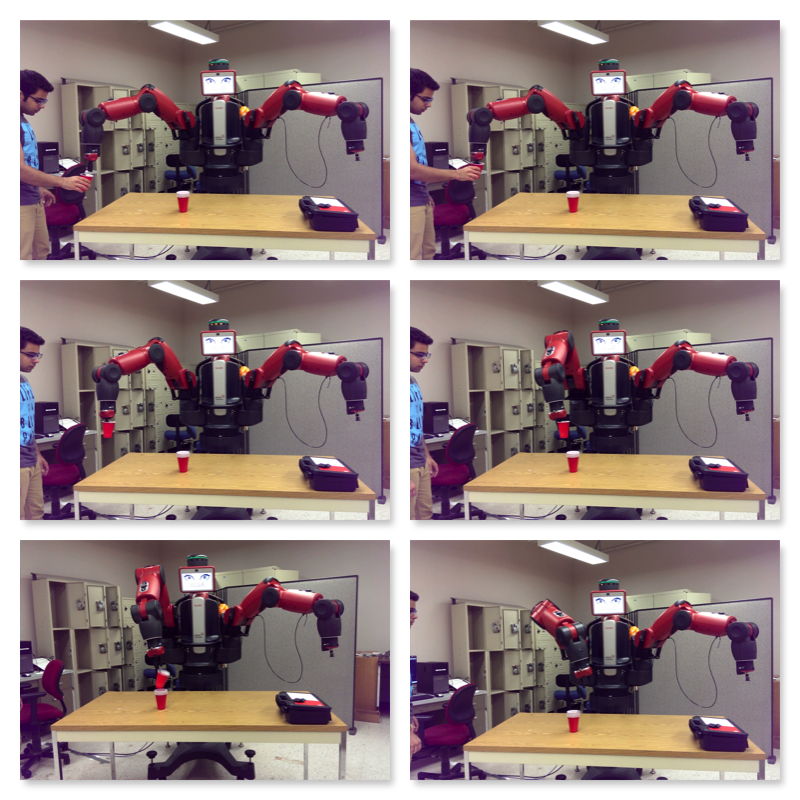}
\end{center}
   \caption{Robot performing learned stacking cup task}
\label{fig:demo_cup}
\end{figure}

\begin{figure}
\begin{center}
\includegraphics[width=0.9\linewidth]{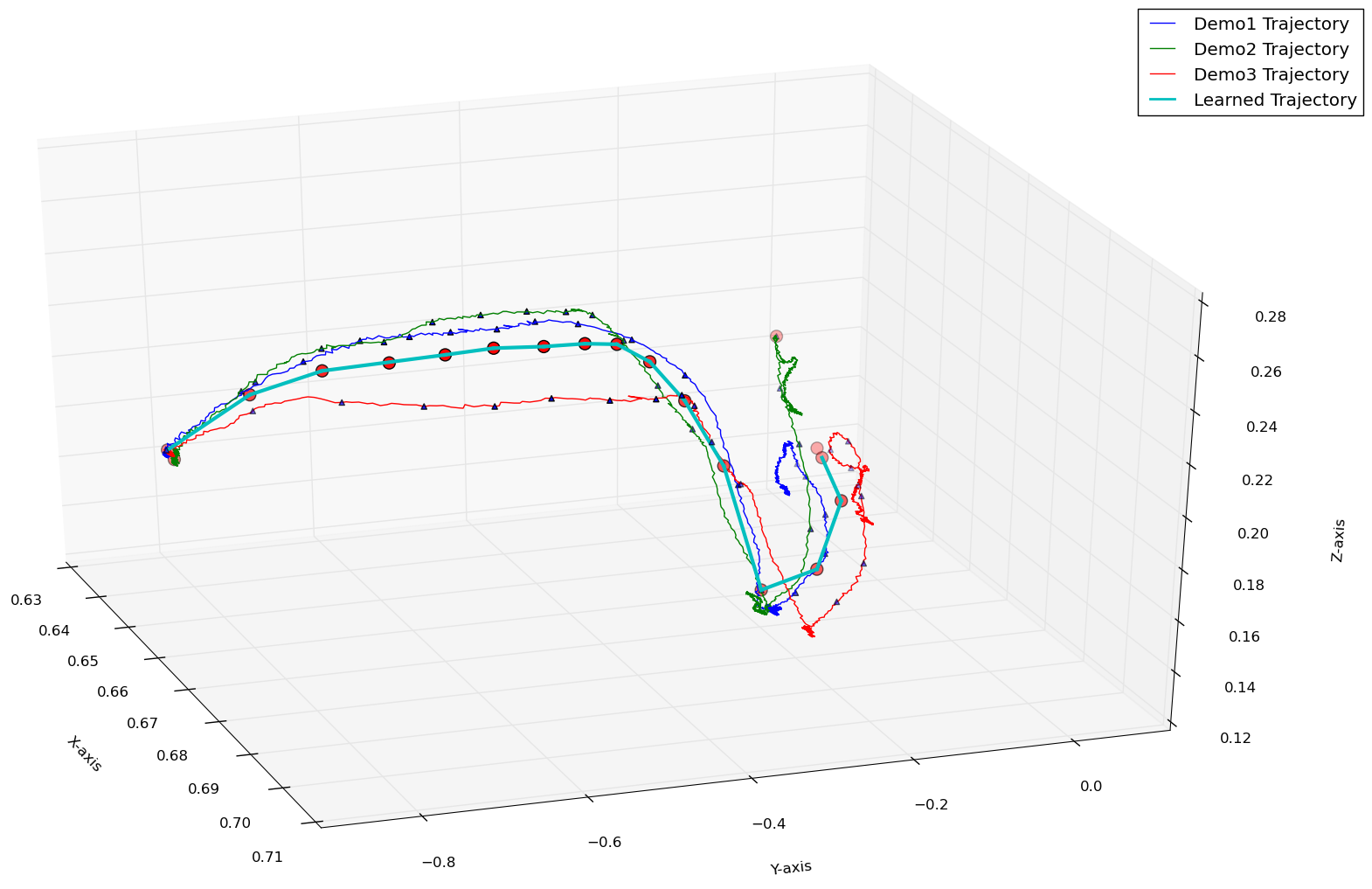}
\end{center}
   \caption{Demonstrated and learned trajectory for stacking cup task}
\label{fig:cup_plot}
\end{figure}

The robot was able to reproduce the task successfully. This task involved learning of precise motion of the manipulator to precisely drop the cup on top of the stacked cups. Figure \ref{fig:cup_plot} shows the manipulator position for the demonstrated task and the learned trajectory.  The small triangles on the trajectories marks the key-points extracted from that demonstration, and the red circles show the centroids of the clusters formed. In total, 19 clusters were formed and they acted as states for the HMM. Numbers from 0-18 named these centroids. The cyan line shows the generalized trajectory. This plot only shows the position of the manipulator only. They complete learning process involved the learning of the orientation of the manipulator as well. Robot performing the learned stacking cup task can be seen in figure \ref{fig:demo_cup}.

\subsection{Pick and place task}

The second task is a pick and place task, in which the robot picks up a block of wood and places it at some other spot and then pick the block from that spot and places it back to the original spot. This is a much more complex task than the previous task. Table 4 shows the data statistics for this task.

\begin{figure}
\begin{center}
\includegraphics[width=0.9\linewidth]{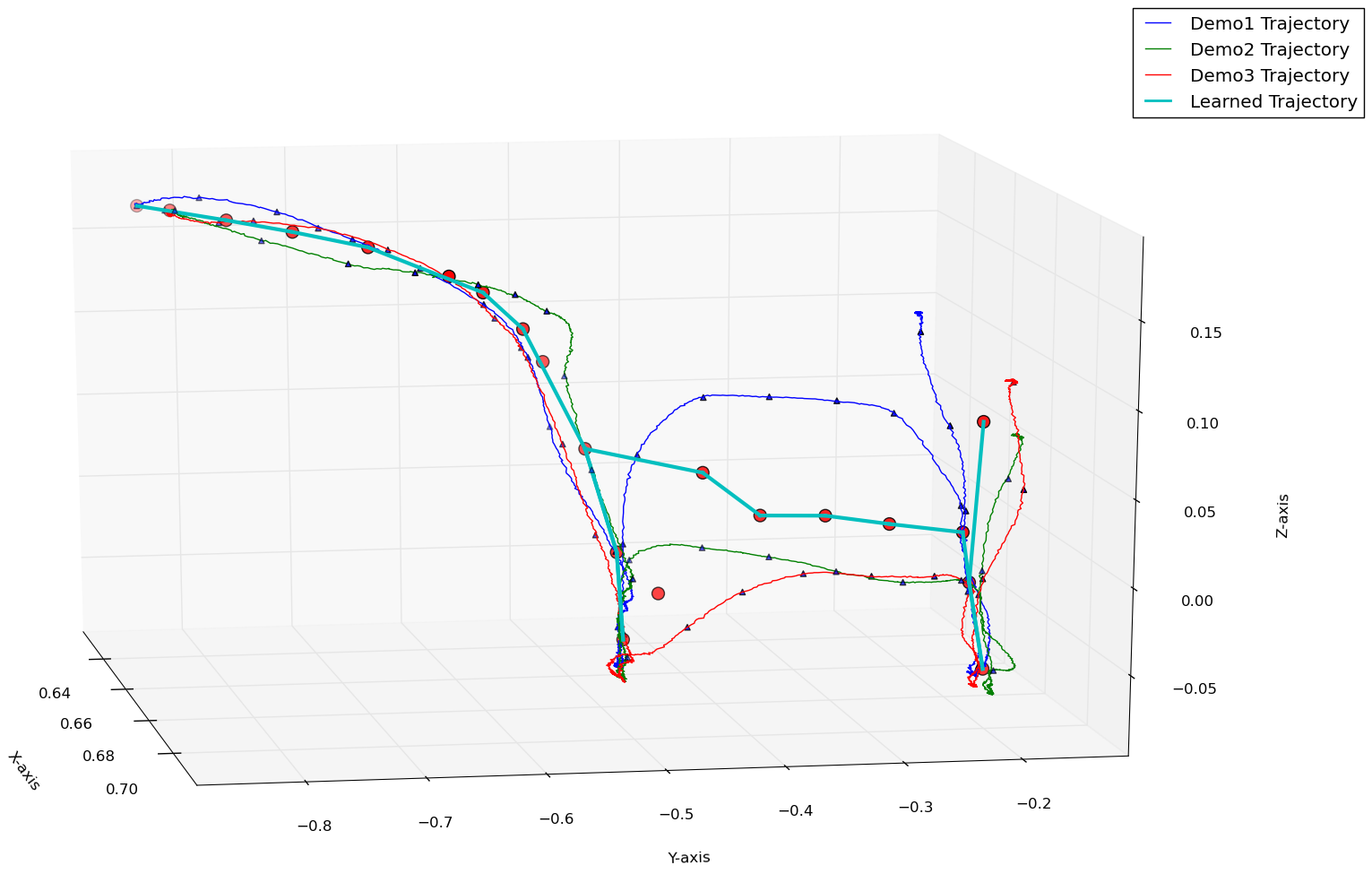}
\end{center}
   \caption{Demonstrated and learned trajectory for pick and place task}
\label{fig:pp_plot}
\end{figure}

It can observed in Figure \ref{fig:pp_plot} that the three demonstrations were a lot different from each other, though they were performing the same task. The small triangles on the trajectories marks the key-points extracted from that trajectory, and the red circles show the centroids of the clusters formed. In total, 21 clusters were formed and they acted as states for the HMM. Numbers from 0-20 named these centroids. The cyan line shows the generalized trajectory. Please note that the trajectory in the graph is not the actual trajectory the robot went through. The derived trajectory in the plot is just by connecting the states with linear lines to illustrate the flow of states. The actual trajectory of the robot used JTAS spline interpolation and was much smoother than the one in the plot.

\begin{table}[h]
\centering
\caption{Data statistics for pick and place task}
\label{data_pp}
\begin{tabular}{|c|c|c|}
\hline
Demonstration \# & \# of recorded values & \# of key-points extracted \\
\hline
Demonstration  1     & 31510 & 24  \\
Demonstration  2     & 28593 & 21   \\
Demonstration 3      & 27025 & 21 \\
\hline
\end{tabular}
\end{table}

In both the experiments, the robot was able to learn the complete task and successfully reproduce the learned task.


\section{DISCUSSION}

In this paper, we presented a framework for RLfD. This framework has three phases: Data collection, Task Learning and Robot Execution. The data collection module collects and stores the eight dimensional time-series data from each demonstration. The Task Learning module then uses this data to learn a model of the task. As we can observe from the experiments, the key-points are extracted from each of the recorded demonstration. This was done to limit the number of observed states for the HMM. As we can observe from the data statistics of the two experiments, the number of extracted key-points is much less than the recorded values. This reduction in the number of states lead to a faster learning of the model. Initially, we tried to learn the model of the task based on the joint angles values, and it avoided the implementation of the forward and inverse kinematics. Then, we implemented FK and IK and used the position and orientation of the manipulator, which made it easier to visualize the trajectories of the task. In addition, the key-points extracted from both the methods were the same in most of the task. Thus, we decided to move forward with the second approach using the position and orientation, as it is easier to plot and visualize the position of the manipulator.

The model learning part of our approach is similar to \cite{Vakanski2012TrajectoryWarping}, but the key point extraction technique we implemented allowed us to precisely extract key-points from the demonstrations. By adjusting the threshold values $\epsilon_1 - \epsilon_{14}$, we were able to fine-tune the key point extraction algorithm to extract only sharp corners from the demonstrated trajectory. The \textit{k}-means clustering technique that we used for clustering the key points allowed us to include all the key points from all the demonstrations performed. Thus, we were able to preserve the key features of the task that were missing from some demonstrations.  Thus, we got more key features of the task as compared to Asfour \etal \cite{Asfour2008ImitationRobots}, where they used the common key points concept, i.e. they just use the key points that are found in all demonstrations. Therefore, the generalized trajectory only includes the features that are present in all demonstrations.
 	
It was experimentally determined that three or four demonstrations for most of the task, tends to produce a generalized trajectory closest to the most probable demonstrated trajectory. The learning of the orientation played a crucial role as well. For many tasks like picking up the wooden block, the orientation of the gripper has to be perfectly aligned with the object to grab the object from the right place. The model learning also involved learning of the gripper states. The model learned should not only learn whether the gripper should be closed or open, but also the width it should open or close to grasp or release an object. This ensures that the robot does not break delicate objects by learning the grasping width of the gripper during the task.
    

\section{CONCLUSIONS}

In this paper, a novel RLfD framework was introduced, which used HMM to model the trajectory skill demonstrated to a collaborative robot. Our results show that the robot can collaborate with the human teacher and learn variety of trajectory based task in short time and reproduce the task while working with a human user.

In future, other collaborative robots can be trained to learn complex task using the developed RLfD framework. With the advancements in the field of image processing and visual servoing, RLfD can be made more robust by allowing the robot to learn the position and orientation of the task relevant objects in addition to the position and orientation of the manipulator. This way, the robot will be able to adapt to the changing environment, as it will learn the environment parameters like distance from task relevant objects. This will enable the robot to learn a task and then reproduce it in a dynamic environment.

\addtolength{\textheight}{-12cm}   




{\small
\bibliographystyle{ieeetr}
\bibliography{mybib}
}

\end{document}